\documentclass[letterpaper, 10 pt, conference]{ieeeconf}  

\IEEEoverridecommandlockouts                              

\usepackage{pifont}
\usepackage{multirow}
\usepackage{booktabs}
\usepackage{graphicx}
\usepackage{makecell}
\usepackage{url}
\usepackage{algorithm}
\usepackage{algorithmic}
\usepackage[T1]{fontenc}
\usepackage[table]{xcolor}
\usepackage{tabularx}
\usepackage{bbding}
\usepackage{hyperref}
\usepackage{subcaption}

\title{\LARGE \bf
PeriGuru: A Peripheral Robotic Mobile App Operation Assistant
based on
GUI Image Understanding and Prompting with LLM
}

\author{Kelin Fu, Yang Tian and Kaigui Bian
\thanks{The authors are from School of Computer Science, Peking University, China. Correspondence to: {\tt\small bkg@pku.edu.cn}}
}
\begin{document}

\maketitle
\thispagestyle{empty}
\pagestyle{empty}

\begin{abstract}

Smartphones have significantly enhanced our daily learning, communication, and entertainment, becoming an essential component of modern life. However, certain populations, including the elderly and individuals with disabilities, encounter challenges in utilizing smartphones, thus necessitating mobile app operation assistants, a.k.a. mobile app agent. With considerations for privacy, permissions, and cross-platform compatibility issues, we endeavor to devise and develop \emph{PeriGuru} in this work, a peripheral robotic mobile app operation assistant based on GUI image understanding and prompting with Large Language Model (LLM). PeriGuru leverages a suite of computer vision techniques to analyze GUI screenshot images and employs LLM to inform action decisions, which are then executed by robotic arms. PeriGuru achieves a success rate of 81.94\% on the test task set, which surpasses by more than double the method without PeriGuru's GUI image interpreting and prompting design. Our code is available on \href{https://github.com/Z2sJ4t/PeriGuru}{https://github.com/Z2sJ4t/PeriGuru}.

\end{abstract}

\section{Introduction}

Smartphones have become an integral part of daily life, serving as tools for reading, learning, socializing, and shopping. As anticipated in 2024, an estimated 4.88 billion individuals possessing smartphones, dedicating an average of 4.37 hours daily to their usage~\cite{smartphone_usage}. Moreover, the dynamic market of mobile apps illustrates its vitality as close to 1000 novel apps are introduced daily on Google Play~\cite{googleplay}.

However, not all users find mobile apps easy to navigate. Seniors, for instance, often encounter barriers, including difficulty in locating small text and/or icons, confusion with user interfaces and functionalities, and compounded by a lack of familiarity with app operations~\cite{elderly_smartphone}. Furthermore, global statistics from the World Health Organization (WHO) exhibit that approaching 2.2 billion people worldwide experience near or distant vision impairment~\cite{WHO}. Previous studies have revealed that the measures for visually impaired individuals in mobile apps are insufficient, with 70\% of apps lacking accessibility tags~\cite{LabelDroid}. Thus, an intelligent app operation agent has the potential to play an assistant role and facilitate the lives of this demographic. Simultaneously, even adults without perceptual or cognitive impairments can immensely benefit from these assistants, whether to bypass learning new app operations or expedite repetitious processes.


While numerous studies have explored automated app operations within the domain of software testing, the prevailing methodologies often rely on tools that capture screen position information at the software level, such as UIAutomator~\cite{uiautomator}. Yet, in the realm of user assistants, we advocate for a purely peripheral approach. Our investigation has identified several key reasons for this preference:

\begin{itemize}
    \item \textbf{Privacy concerns of built-in software on phones}: Mobile phones frequently contain substantial personal data. Generally, users would be resistant to granting software assistants the rights to oversee screens, manipulate files, and control software on their devices. A purely peripheral solution emerges as a safer alternative.
    \item \textbf{Permission constraints of voice control in mobile apps}: Dependence on software for screen information and mobile phone manipulation demands elevated permissions. Many voice-control assistants are restricted from operating within third-party apps, due to these permission constraints. This is particularly challenging for user groups such as the elderly, as they often struggle with high-permission operating systems, fearing the potential for irreversible misoperations.
    \item \textbf{Compatibility/applicability across platforms}: Although Android and iOS dominate the global mobile operating system landscape, emerging platforms such as 
    Surface Duo~\cite{surfaceduo} and HarmonyOS~\cite{HarmonyOS} 
    are gaining traction. Additionally, some manufacturers have crafted simplified operating systems targeted at seniors. Software-based assistants struggle to maintain compatibility across this diverse array of systems. 
\end{itemize}

Hence, we seek to develop an embodied agent as a purely-peripheral mobile app assistant. It does not require any data access, voice-control, or any other permission from the mobile OS/apps, being naturally cross-platform compatible.

In this paper, we present our embodied agent, \emph{PeriGuru}. Leveraging Computer Vision (CV) methodologies, PeriGuru interprets screenshot images to formulate prompts and further exploits large language models (LLMs) for effective decision-making. This approach enables it to autonomously execute a range of app operations on behalf of users, with no need to invoke any software APIs of the smartphone. PeriGuru represents a significant advancement in the development of mobile app agent based on image understanding. Upon evaluation, PeriGuru demonstrates an impressive performance, achieving a total plan success rate of 89.71\% and an execution success rate of 81.94\%. These results underscore PeriGuru's effectiveness in facilitating app operations for users, thereby enhancing accessibility and usability.

\section{Limitations of Generalized Large Language Models}
\label{sec:motivation}

\begin{figure}[tb]
\centering
\includegraphics[width=0.99\linewidth]{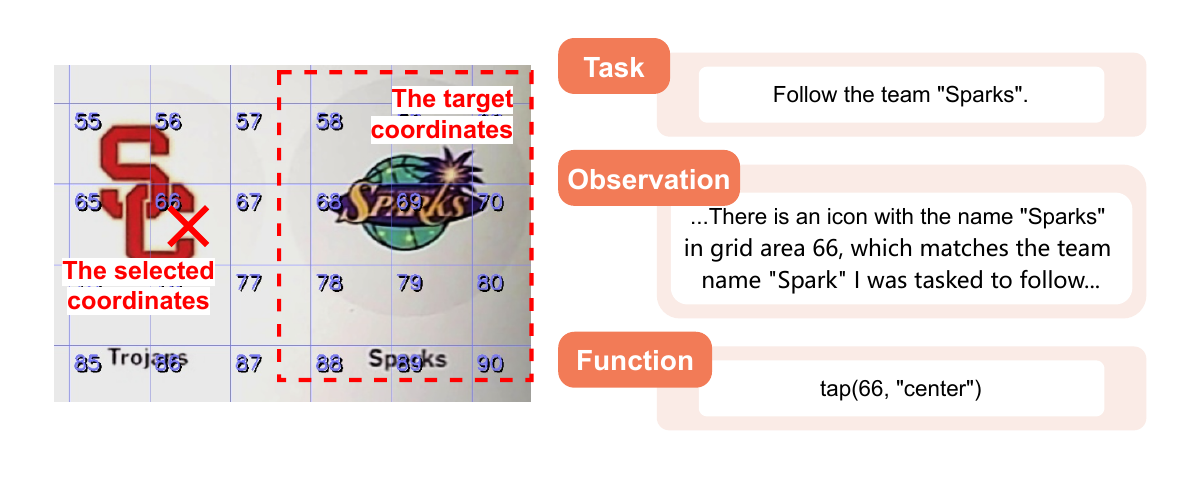}
\caption{A showcase of coordinate misunderstanding by AppAgent (grid). On the left is a portion of the screenshot with added grids, and on the right is a part of the output provided by \textit{gpt-4-vison-preview}. The task is to follow the team "Sparks", whcih means the agent should tap on the region that encompasses the team's logo, as indicated by the red dashed box on the left. However, the agent's actual selection is marked by the red cross, which deviates from the intended target area and, consequently, fails to accomplish the task as prescribed.
}
\label{fig:grid_mode}
\end{figure}

\begin{table}[tb]
  \centering
  \begin{tabular}{cc}
    \hline 
    \rowcolor{blue!20} Failure Reason & Proportion \\ \hline
    \textit{Missing important icons} & 36.67\% \\ 
    \rowcolor{gray!20} \textit{Wrong decison} & 30.00\%\\
    \textit{Wrong coordinates} & 13.33\%\\
    \rowcolor{gray!20} \textit{Misunderstanding ordinal numbers} & 10.00\%\\
    \textit{Misunderstanding icon meanings} & 10.00\% \\
    \bottomrule 
   \end{tabular}
    \caption{The most common reason for tasks that AppAgent (grid) fails to execute.
    }
    \label{tab:fail_reason}
    \vspace{-5mm}
\end{table}

The evolution of Large Language Models (LLMs) has been remarkable, with models like GPT-4V~\cite{gpt4v} setting new benchmarks by demonstrating the capacity of LLMs to process and interpret visual data. This capability significantly extends the reasoning and decision-making potential of LLMs, positioning them as increasingly viable solutions for embodied intelligent agents.

However, without precise text prompting, generalized LLMs display several shortcomings. For a comprehensive analysis, we conducted an experiment comprising 72 task sets using the \texttt{grid} mode of AppAgent~\cite{APPAgent}, a state-of-the-art LLM-based app operation agent. In this mode, the identification of GUI elements is only achieved through the coordinate representation of the gridded image, without utilizing the software GUI metadata typically required for AppAgent. An example of image gridding and corresponding LLM output is shown in Fig.~\ref{fig:grid_mode}. From the 30 tasks that ended in failure, we derived several primary sources of errors\footnote{\textit{Wrong decision} implies that the agent selected an action completely unrelated to the task at hand. \textit{Misunderstanding ordinal numbers} indicates that an incorrect ordinal number was chosen for tasks involving an order of elements. \textit{Wrong coordinates} suggests that the provided coordinates did not correspond to the GUI elements selected for interaction.}, as documented in Table~\ref{tab:fail_reason}. The results highlight the key challenges in app operation tasks for agents:

\begin{itemize}
    \item \textbf{Icon recognition failure}: Icons pose a significant challenge in app operation tasks. Enhancing the agent's capability to correctly identify and interpret these visual elements is instrumental in the successful execution of app operation tasks.
    \item \textbf{Coordinate and ordinal misunderstanding}: LLMs exhibit difficulties in comprehending coordinates and ordinal numbers. In fact, of the 42 tasks successfully planned, 38.10\% were improperly carried out by the robotic arm due to inaccurate coordinates.
\end{itemize}

In light of these insights, PeriGuru incorporates a composite CV approach to locate screen elements, discern layout configurations, element sequences, and the significance of icons.

\begin{figure*}[htbp]
\centering
\includegraphics[width=0.99\textwidth]{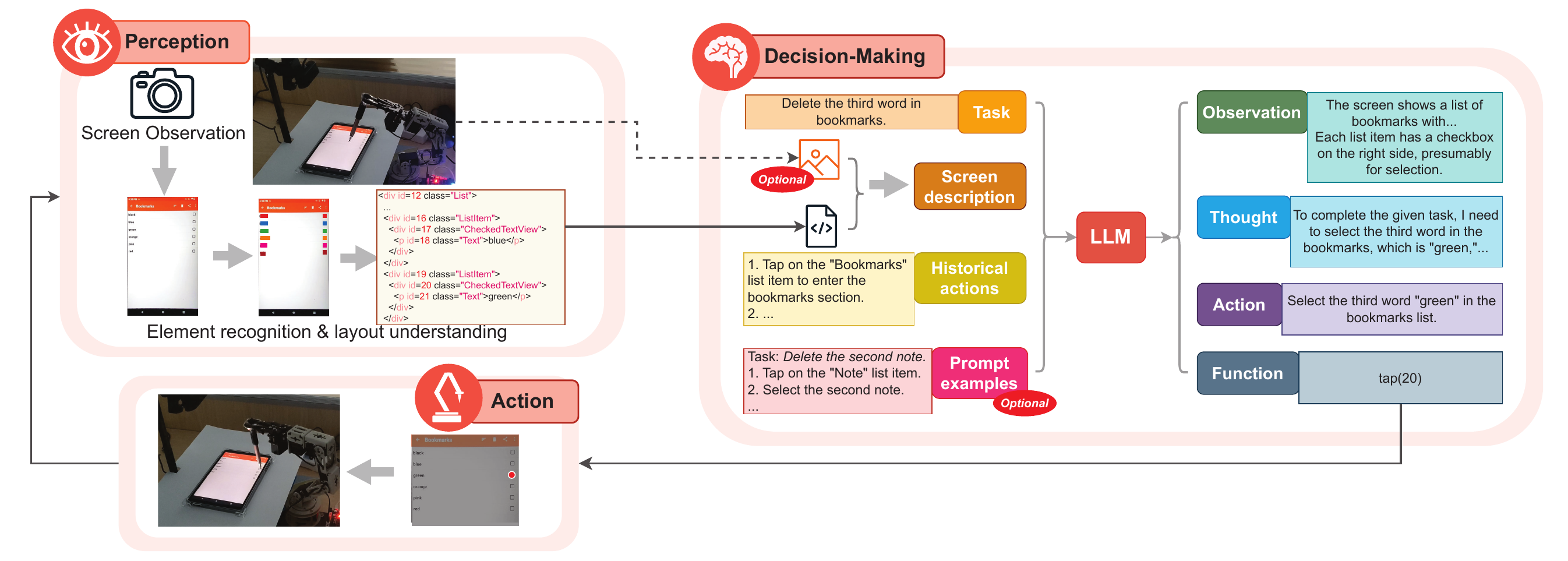}
\caption{PeriGuru's system overview.}
\label{fig:system}
\vspace{-0.2in}
\end{figure*}

\section{Method}

Fig.~\ref{fig:system} showcases the overall architecture of PeriGuru. 
The process comprises three distinct stages as follows:

\textbf{Perception stage (Sec.\ref{sec:perception_stage}).} Initially, a high-speed camera captures a raw image, which undergoes fundamental CV processing to extract a screenshot. PeriGuru then performs element and layout recognition to formulate a detailed screenshot description. This description, along with the labeled screenshot, is forwarded as screen observation data to the subsequent stage.

\textbf{Decision-making stage (Sec.\ref{sec:dm_stage}).} Leveraging the screen observation information, along with a record of historical actions and analogous task instances, PeriGuru crafts prompts to interface with a LLM to obtain the subsequent action decision in the form of a function call. 

\textbf{Action stage (Sec.\ref{sec:action_stage}).} PeriGuru then translates the decision formulated in the preceding stage into a series of robot actions. Additionally, identifiable errors are integrated back into observation information, enhancing the framework's capacity for error-aware decision-making.

\subsection{Perception: GUI Layout Understanding}
\label{sec:perception_stage}

\begin{figure*}[htbp]
\centering
\includegraphics[width=0.99\textwidth]{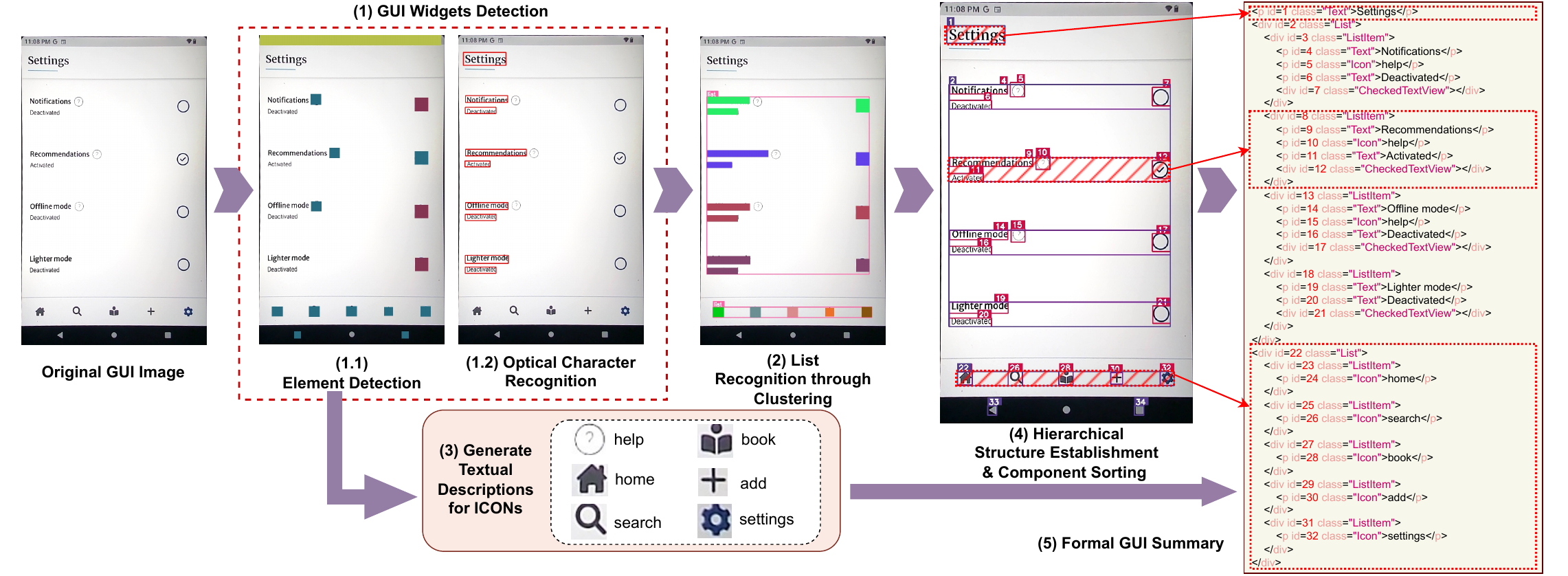}
\caption{PeriGuru's GUI image process architecture.}
\label{fig:gui_process}
\vspace{-0.2in}
\end{figure*}

As shown in Fig.~\ref{fig:gui_process}, PeriGuru adopts a purely image-based methodology for identifying the components, layout, and interaction logic of mobile GUIs. This approach bypasses the need for access to any underlying code or GUI metadata of the apps. The recognition process encompasses several key components.

\begin{figure}[tb]
\centering
\includegraphics[width=0.99\linewidth]{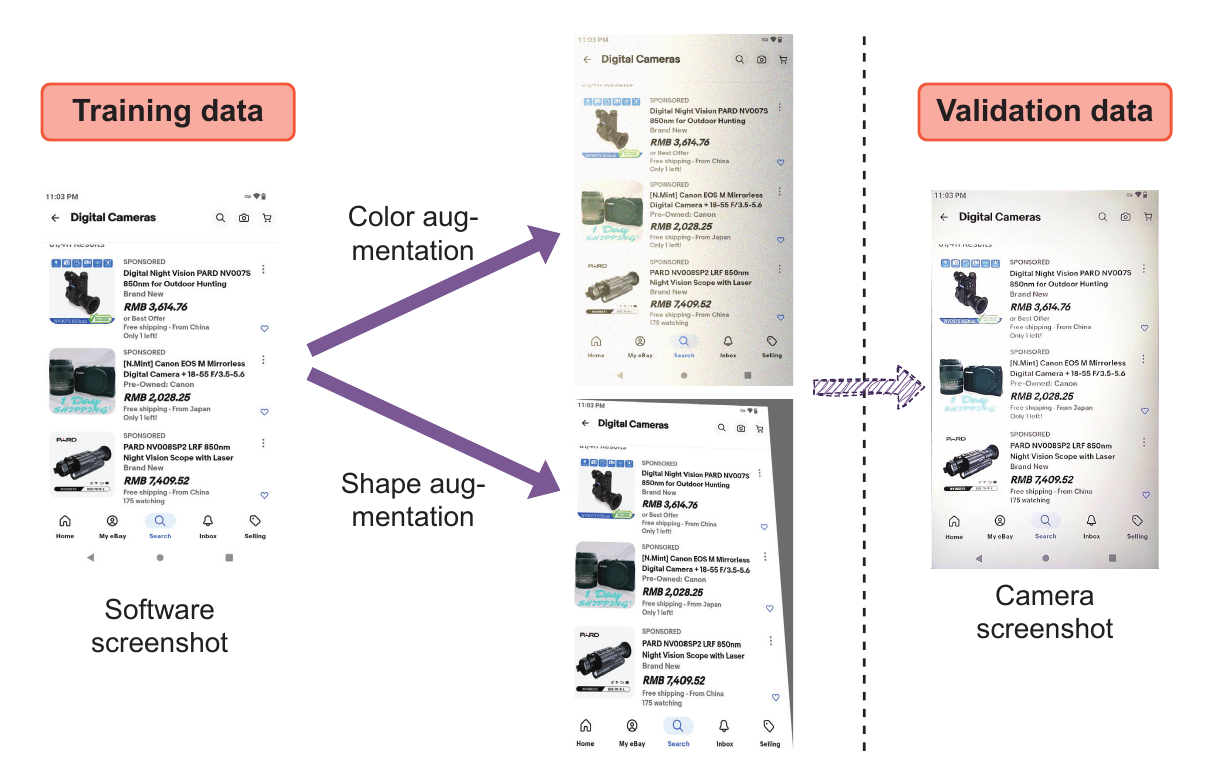}
\caption{Augmentation of color and shape make the training set data collected through software screenshots closer to the training set captured by the camera, and thus enhance the robustness of the object detection model.
}
\label{fig:augmentation}
\vspace{-0.2in}
\end{figure}

\textbf{GUI widgets detection.} Compared to images captured by cameras, software screenshots are much easier to collect and thus are more conducive to the assembly of large-scale training datasets. However, in practical application environments, images obtained through cameras are frequently subject to variations in ambient lighting and camera angles, which can compromise the image quality when compared to those derived from software screenshots. To address the discrepancy between the quality of training data from software screenshots and the validation data from camera captures, we implemented a data augmentation strategy aimed at bolstering the robustness of our detection model. We augmented data in terms of both color and shape by adding uneven light masks and noise, and applying rotations and perspective deformations, as depicted in Fig.~\ref{fig:augmentation}. This can make the distribution of the training set more similar to the validation set. In Sec.~\ref{sec:exp_res}, we verified the performance improvement of this enhancement for GUI widgets detection.

The dataset we leveraged for the detection of non-textual GUI components is VINS~\cite{VINS}. It provides a more accurate re-annotation of the well-established GUI image dataset RICO~\cite{RICO}, enriched with nearly 2000 additional sets of screenshot data from Android and Apple apps. This dataset enables the precise recognition of significant GUI components, including icons, images, checkboxes, text input fields, pop-ups, buttons, and sidebar menus. As for the setup of object detection algorithm, We opted for the contemporary YOLOv5 framework~\cite{YOLOv5} which showed impressive results in terms of both detection accuracy and response speed.

Complementarily, to identify and process texts---a key GUI feature signalling function---we resort to mature optical character recognition (OCR) services, such as Google OCR~\cite{google_ocr} or Baidu OCR~\cite{baidu_ocr}. 

We have also ensured a modular design for our code, granting seamless replacements of GUI widget detection methods and other GUI perception pipeline components with alternative algorithms. Consequently, this section can deploy varying object detection algorithms, regardless of their foundations in traditional CV schemes or learning-based approaches, as well as different OCR interfaces.

\textbf{List recognition through clustering.} GUI interfaces are not merely random amalgamations of independent widgets. Rather, they are thoughtfully organized with widgets of similar semantics grouped together to aid human comprehension of interface functionality. This approach is supported by research in psychology and biological vision~\cite{Gestalt}, and it continues to be widely implemented in the field of GUI design. Consequently, the ability to correctly discern this list-like structure is crucial for our agent to understand the functionality of GUIs.

Building upon previous research~\cite{UIED2}, we adopt a similar methodology to identify list structures. We utilize the DBSCAN clustering algorithm~\cite{DBSCAN}---implementing it on the spatial distributions, such as widget's area, coordinates, and inter-widget gaps. Significantly, this clustering process also allows us to rectify any missed or incorrectly-detected widgets resulting from the previous stage.

\textbf{Generate textual descriptions for icons.} Beyond text, icons also play an important role in conveying functionality information within a GUI interface. Icons are frequently crafted following a consistent design language, enabling experienced users to discern their meanings without relying on accompanying text. For instance, a magnifying glass typically symbolizes ``search'', while a left-facing arrow indicates ``back''. As a result, assuming a level of user familiarity, not all icons are complemented with textual annotations in certain GUI designs.

Given the significance of interpreting icons for fully understanding GUI functionality, we employ the state-of-the-art icon label generation framework, LabelDroid~\cite{LabelDroid} to generate textual descriptions for all common icons within the interface. These annotations are appended to the final GUI summary, enhancing PeriGuru's comprehension of GUI functionality.

\textbf{Hierachical structure establishment and component sorting.} In this phase, we establish the parent-child relationship between screen components based on intersection over union (IoU) calculations, which facilitates the construction of a tree-like hierarchical structure. Following this, we organize all components based on the natural human eye reading pattern using the widely accepted recursive XY-cut page segmentation algorithm~\cite{xycut1}.

\textbf{Output formal GUI summary.} Given that the majority of LLMs' training data originates from raw web pages, an HTML-like format is highly conducive to conveying screen observation information to LLMs~\cite{prompt_need}. With this format, each screen element includes these essential attributes:
\begin{itemize}
    \item \texttt{id}: The component's unique resource id.
    \item \texttt{class}: The category of the component, which typically encompasses types such as \texttt{Text}, \texttt{Image}, \texttt{Icon}, \texttt{CheckedTextView}, etc.
    \item \texttt{content}: Additional descriptions of components, such as textual content and icon explanations.
\end{itemize}

\subsection{Decision-Making: LLM Agent Establishment}
\label{sec:dm_stage}

The domain of utilizing LLMs for decision-making encompasses two principal methods: one is to generate long-term action plans as seen in works such as ProgPrompt~\cite{ProgPrompt} and LLM-Planner~\cite{LLM_planner}; the other involves selecting the immediate, most relevant action to the task, epitomized by the work on discussing robotic affordances, SayCan~\cite{SayCan}.

In our scenario, the limited observational scope of operating app tasks makes it challenging to directly create long-term action plans. As a result, we are obliged to design prompts for the LLM to choose a single action step. The design of GURU's LLM prompt is designed as follows.
\begin{itemize}
    \item \textbf{Task}: The description of mobile app operation task that should be executed.
    \item \textbf{Screen description}: This is an HTML-like representation generated during the perception stage, encapsulating the visual elements of the GUI interface. For LLMs that accommodate multimodal inputs, a screenshot image is selectively integrated alongside the description.
    \item \textbf{Historical actions}: These are the actions previously undertaken by the agent, complemented by feedback on their actual execution. For instance, if a keyboard is absent when required for a \texttt{Text()} action, the corresponding error information will be reported in this section.
    \item \textbf{Prompt examples}(optional): By using a K-Nearest Neighbors (KNN) search to find the most analogous task in the case database and adding the execution steps of it into the prompts, the LLM agent can draw guided reference from case studies and make better decisions.
\end{itemize}

The output from the LLM is required through prompts to provide a summary of the current state, reflections on the forthcoming decision, a natural language explanation of the subsequent action, and ultimately the requisite function call, which is exhibited in detail in Fig.~\ref{fig:system}. Aligning with previous study~\cite{CoT}, this approach facilitates the generation of a ``Chain of Thought (CoT)" by the LLM, thereby enhancing its decision-making capabilities.

\subsection{Action: Action Space Design}
\label{sec:action_stage}

Taking into account the capabilities of the robotic arm and the common actions ordinarily undertaken by human users, we have delineated the following primitive actions, encapsulated as a functional construct: 

\begin{itemize}
    \item \texttt{Tap(id: int)}: This function simulates a tap on a UI element identified by its unique ID.
    \item \texttt{Long\_press(id: int)}: This function emulates a long press on a UI element identified by its unique ID.
    \item \texttt{Text(text: str)}: To expedite text input, this function process, rather than emulating a series of \texttt{Tap()} operations for letter keys, we leverage this function to perform a sequence of typing actions when the keyboard is invoked.
    \item \texttt{Scroll(direction: str)}: This function executes a scrolling gesture on the screen, with \texttt{direction} specifying the desired scrolling direction.
    \item \texttt{Back()}: To facilitate the agent's navigation away from irrelevant interfaces, this function is established as a distinct primitive action, rather than utilizing the \texttt{Tap()} function on a back button.
    \item \texttt{Finish()}: This function signifies the completion of the given task, requiring no further action from the robotic arm.
\end{itemize}

\begin{figure*}[tb]
    \centering
    \includegraphics[width=0.99\textwidth]{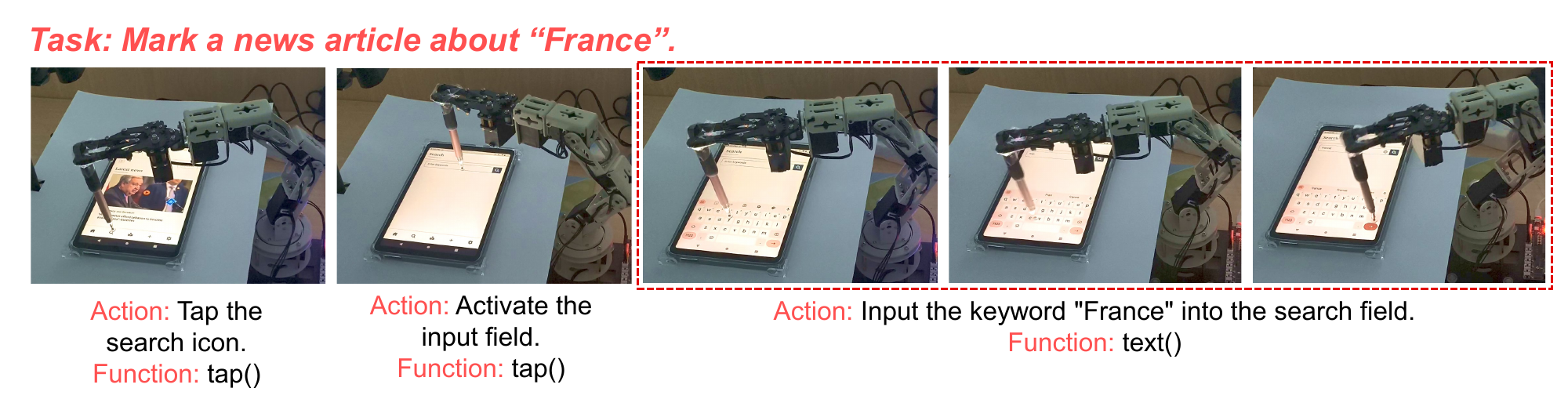}
    \caption{An example of PeriGuru's robotic arm testbed. It recognizes the GUI elements such as the search icon and keyboard, to organize the robotic arm to execute a sequence of actions that align with decisions rendered by LLM.}
    \label{fig:demo}
    \vspace{-0.1in}
\end{figure*}

\section{Experiments}

\subsection{Methodology}

\textbf{Task dataset.} To construct a task dataset, we firstly selected 10 apps spanning four prevalent domains: news, social media, shopping and learning, and create a series of tasks manually. For each app, we manually designed a series of tasks. These tasks encompass the utilization of core app functionalities, such as searching for news articles in news apps and selecting products in shopping apps, as well as common user interactions, including adjusting settings and navigating to a certain interface. This initial task set comprises 36 single-step tasks and an equal number of multi-step tasks. For the multi-step tasks, the average optimal step count for multi-step tasks stands at 3.528.

Then, to expand our task dataset, we extracted a collection of screenshots from the RICO~\cite{RICO} dataset. The RICO dataset offers valuable user interaction traces and UI hierarchy files. By utilizing these resources, we leveraged LLM to generate an additional 64 task descriptions. These generated tasks will are henceforth referred to as RICO tasks.

\textbf{Metrics.} For performance comparison, we employed three distinct metrics:
\begin{itemize}
    \item \textbf{Plan Successful Rate (Plan SR)}: This metric is derived from simulation operations performed through software interfaces. After the mobile app agent provides the type and coordinates of the next operation, we perform it on the proximate interactive element found by GUI metadata acquired through UIAutomator~\cite{uiautomator}. The success rate in this context is referred to as the plan SR.
    \item \textbf{Execution Successful Rate (Execution SR)}: To measure the success rate in practical operation, we established a testbed with Yahobom DOFBOT SE~\cite{dofbot} robotic arm, which is shown in Fig.\ref{fig:demo}. Within this setup, we assessed each method's execution SR.
    \item \textbf{Average Steps}: This metric represents the average step count for all successfully executed tasks, serving as a reference for efficiency.
\end{itemize}

\textbf{Baseline.} We take the \textit{grid} version of AppAgent as the baseline model. A detailed explanation of it can refer to Sec.\ref{sec:motivation}. Furthermore, we conducted an investigation into the differences in performance when relying solely on GUI element detection and icon meaning recognition, as opposed to developing a hierarchical screen description, denoted as \textit{element labels} in Table.\ref{tab:result} or \textit{label} in Fig.\ref{fig:result}, or utilizing a comprehensive PeriGuru prompt, referred to as the \textit{HTML-like document} in Table.\ref{tab:result} or \textit{document} in Fig.\ref{fig:result}.

Post the refinement of text prompts, the original screenshot image may become dispensable for LLMs. Models that operate without images offer the advantages of accelerated processing speeds and reduced token consumption, which are potentially desirable options for the construction of mobile app agents. Consequently, we tested two LLM backbones in parallel: the multimodal model \texttt{gpt-4-vision-preview} that incorporates images, and the image-less model \texttt{gpt-4o}. Both models are provided by OpenAI~\cite{openai}.

\subsection{Results}
\label{sec:exp_res}
\begin{table*}[t]
    \belowrulesep=0pt
    \aboverulesep=0pt
  \centering
  \resizebox{0.95\textwidth}{!}{
  \begin{tabular}{c | c c c| c c c | c c |c}
        \hline 
        \multirow{2.5}{*}{Framework} & \multirow{2.5}{*}{Prompt format} & \multirow{2.5}{*}{LLM backbone} & \multirow{2.5}{*}{Use image} & \multicolumn{3}{c}{Plan SR} & \multicolumn{2}{c}{\makecell{Execution \\ SR}}  & \makecell{Average \\steps} \\
        \cmidrule{5-10} & & &  & multi & single & RICO & multi & single & multi \\ \toprule
        AppAgent (grid) & grid & gpt-4-vision-preview & \Checkmark & 0.611 & 0.556 & 0.875 & 0.333 & 0.389 & 5.09 \\ \hline
        \multirow{4.5}{*}{PeriGuru} & \multirow{2}{*}{element labels} & \cellcolor{gray!20}gpt-4-vision-preview & \cellcolor{gray!20}\Checkmark &  \cellcolor{gray!20}\underline{\textit{\textbf{0.861}}} & \cellcolor{gray!20}0.833 & \cellcolor{gray!20}0.953 & \cellcolor{gray!20}\underline{\textit{\textbf{0.806}}} & \cellcolor{gray!20}0.806 & \cellcolor{gray!20}\underline{\textit{\textbf{4.17}}}  \\
        & & gpt-4o & \XSolidBrush & 0.722 & 0.694 & 0.859 & 0.667 & 0.694 & 5.23 \\ \cmidrule{2-10}
        & \multirow{2}{*}{\makecell{HTML-like \\document}} & \cellcolor{gray!20}gpt-4-vision-preview &\cellcolor{gray!20}\Checkmark & \cellcolor{gray!20}0.833 & \cellcolor{gray!20}\underline{\textit{\textbf{0.861}}} & \cellcolor{gray!20}\underline{\textit{\textbf{0.969}}} & \cellcolor{gray!20}\underline{\textit{\textbf{0.806}}} & \cellcolor{gray!20}0.833 & \cellcolor{gray!20}4.19 \\
        & & gpt-4o & \XSolidBrush & 0.806 & \underline{\textit{\textbf{0.861}}} & 0.938 & 0.722 & \underline{\textit{\textbf{0.861}}} & 5.38 \\
        \bottomrule
        
    \end{tabular}
    }
    \caption{Performance evaluation results.}
    \label{tab:result}
    \vspace{-3mm}
\end{table*}

\begin{figure}
    \centering
    \includegraphics[width=\linewidth]{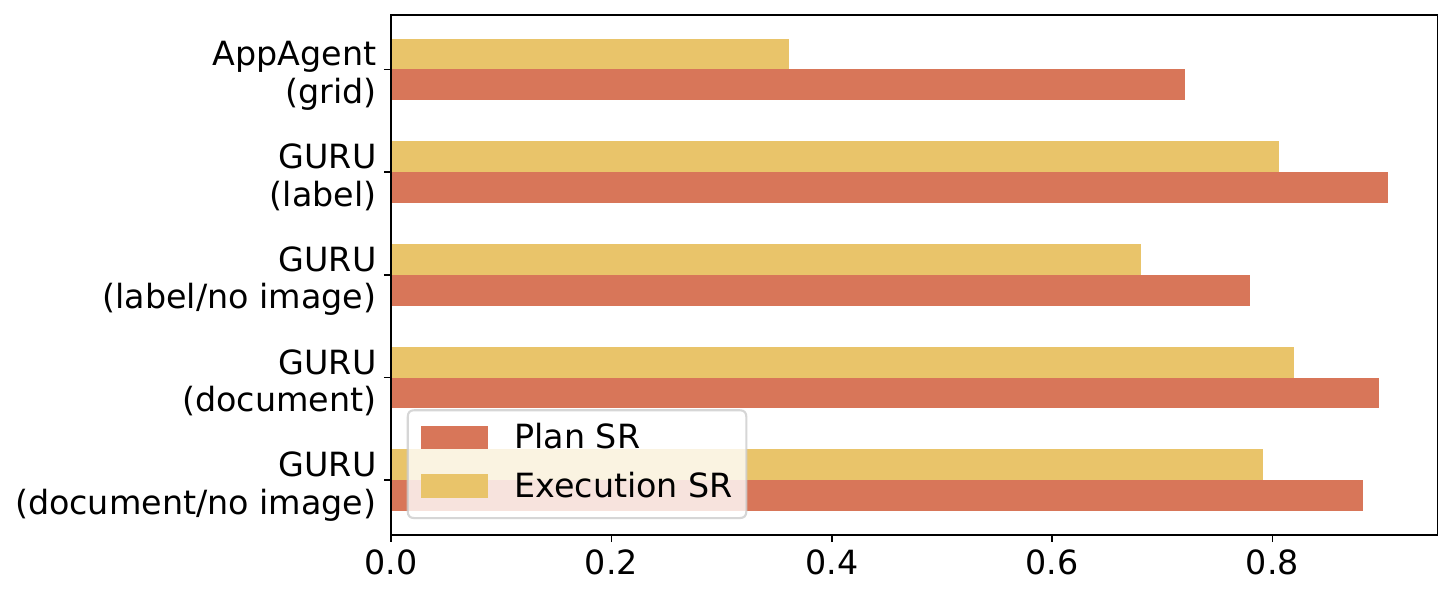}
    \caption{Overall performance of different frameworks. Detailed data is listed in Table~\ref{tab:result}.}
    \label{fig:result}
    \vspace{-0.2in}
\end{figure}
\begin{table*}[t]
    \belowrulesep=0pt
    \aboverulesep=0pt
  \centering
  \resizebox{0.8\textwidth}{!}{
  \begin{tabular}{c | c c c | c c c | c c c}
        \hline 
        IoU threshold & \multicolumn{3}{c}{0.5} & \multicolumn{3}{c}{0.75}  & \multicolumn{3}{c}{0.95} \\ \hline
        Method & Prec. & Recall & F1 & Prec. & Recall & F1 & Prec. & Recall & F1 \\ \toprule
        Basline YOLOv5 & 0.746 & 0.85 & 0.795 & 0.745 & 0.800 & 0.772 & 0.478 & 0.702 & 0.569\\
        \rowcolor{gray!20}Shape aug. & 0.769 & 0.876 & 0.819 & 0.749 & 0.876 & 0.808 & 0.534 & 0.715 & 0.611\\
        Color aug. & 0.795 & 0.844 & 0.819 & 0.785 & 0.844 & \underline{\textit{\textbf{0.813}}} & 0.450 & 0.678 & 0.541\\
        \rowcolor{gray!20}Shape \& color aug. &  0.788 & 0.863 & \underline{\textit{\textbf{0.824}}} & 0.779 & 0.849 & 0.812 & 0.799 & 0.71 & \underline{\textit{\textbf{0.752}}}\\
        \bottomrule
        
    \end{tabular}
    }
    \caption{The effect of data augmentation on improving object detection performance in camera screenshot scenes.}
    \label{tab:augmentation}
    \vspace{-3mm}
\end{table*}

\textbf{Performance enhancement by PeriGuru.} As demonstrated in Table~\ref{tab:result} and Fig.~\ref{fig:result}, PeriGuru has achieved a 24.89\% improvement in plan success rate over AppAgent (grid). This increase underscores the framework's effectiveness in enhancing mobile app agents' comprehension of GUI elements and hierarchical structures. Additionally, the execution success rate has seen a marked increase, rising to 2.3 times the baseline rate, which attests to the significant role of PeriGuru's object detection capabilities in improving the precision of coordinate determination.

\textbf{Comparison of the multimodal and pure text LLM backbones.} When employing multimodal models capable that support images, as evidenced in Table~\ref{tab:result} and Fig.~\ref{fig:result}, the performance of PeriGuru with and without a GUI hierarchical document—--presented in the second and fourth rows, respectively—--does not exhibit significant variation. However, upon comparing the third and fifth rows, it is observed that the incorporation of documents enhances the plan SR by 13.21\% and the execution SR by 16.33\% when using a pure text LLM backbone. Additionally, a comparison between the fourth and fifth rows indicates that similar task success rates can be achieved without the reliance on original screenshot images. The generation of documents renders that original images and multimodal models are no longer absolutely necessary, thereby broadening the range of LLM backbone choices, reducing token consumption, and accelerating response speed.

However, solutions that forgo the use of images often necessitate additional iterations of trial and error and encounter challenges in ascertaining task completion. As demonstrated in Table~\ref{tab:result}, the average number of steps required to complete the task for image-independent solutions—--represented in the third and fifth rows—--is 26.91\% higher than that of solutions which leverage images--—depicted in the second and fourth rows. This observation presents the next potential optimization direction for PeriGuru.

\textbf{The effect of data augmentation on enhancing object detection models.} We further conducted a comparative analysis of the precision, recall, and F1-score for PeriGuru's YOLOv5 object detection models, both with and without the implementation of the data augmentation strategies outlined in Section \ref{sec:perception_stage}. The results are presented in Table~\ref{tab:augmentation}. The findings indicate that data augmentation focused on shape and color, when applied independently, effectively enhances model performance, increasing the average F1 score by 4.78\% and 1.73\%, respectively. Moreover, the model that incorporates both augmentation techniques achieves the most substantial improvement, with an average F1 score increase of 11.80\%, which proves the role of data augmentation strategies in improving model robustness.

\section{Related Work}

\textbf{LLM for Robotics.} Large Language Models (LLMs) are advanced neural networks trained via unsupervised learning with vast parameter sets. The evolution of LLMs, exemplified by GPT-4V~\cite{gpt4v} demonstrates their ability, to process visual data, broadening their role in cognitive tasks and decision-making for embodied agents~\cite{xu2024survey}. Noteworthy contributions in this domain include SayCan~\cite{SayCan}, who have combined task relevance with robotic affordances to inform action selection, deeply impacting LLM applications for embodied agents. ProgPrompt~\cite{ProgPrompt} embodies the benefits of code-based prompts, while Text2Motion~\cite{text2motion} has adeptly generated actionable plans for complex tasks by leveraging varying levels of prompt granularity.

\textbf{Automated mobile APP manipulation.} The growing reliance on mobile applications in daily life boosts the demand for their automated operation. VTest~\cite{vtest} has introduced a robotic testing framework for mobile apps, and another research~\cite{prompt_need} proposed a LLM-based automated testing system. Additionally, there are LLM agents, such as ~\cite{google_CHI, APPAgent}, that serve in the user assistance domain. However, they differ from PeriGuru by requiring software GUI metadata.

\textbf{Image-based GUI Understanding.} In GUI design, software testing, and data monitoring, a lack of GUI metadata often hampers accessibility and label clarity. Studies like ReDraw~\cite{ReDraw}, UIED~\cite{UIED}, Xianyu~\cite{UIED2}, and screen recognition~\cite{screen_recognition} have tackled this with image-based GUI understanding methods.

Meanwhile, icon interpretation is crucial for understanding GUI functionality. In response, researchers have explored various strategies: IconIntent~\cite{iconintent} used key-location-based CV methods to detect sensitive UI elements, while another study~\cite{multimodal_icon} pursued a deep learning-based method for this purpose. Tools like LabelDroid~\cite{LabelDroid} and COALA~\cite{coala} focus on generating concise icon descriptions. All these contributions have significantly informed PeriGuru's development.

\section{Conclusion and Future Work}

In summary, PeriGuru advances assistive robotics by addressing accessibility issues in smartphone use for challenged populations, combining sophisticated computer vision algorithms with the decision-making prowess of LLMs to efficiently operate mobile apps.

However, its focus is mainly on GUI image understanding, with less emphasis on enhancing LLM decision-making, leading to challenges like inaccurate task completion assessment. The untapped potential of LLMs in generating complex robotic actions is a recognized area for PeriGuru's future optimizations.



\bibliographystyle{ieeetr}
\bibliography{reference}

\end{document}